\pdfoutput=1
\documentclass[letterpaper,journal]{IEEEtran}
\usepackage[T1]{fontenc}
\usepackage{amsmath,amssymb,bm}
\usepackage{graphicx,booktabs,multirow,array}
\usepackage{algorithm,algorithmic}
\usepackage{cite}
\usepackage{placeins}
\usepackage{stfloats}
\usepackage[hidelinks]{hyperref}
\graphicspath{{figures/}}
\newcommand{\pcot}{T-PCOT}

\newcommand{\R}{\mathbb{R}}

\begin{document}
\title{Cross-Material Magnetic Core Loss Prediction\\via Operating-Condition Matching\\and Partial Optimal Transport}
\author{\parbox{\textwidth}{\centering
Cong~Yao\textsuperscript{1,$\dagger$}, Yu~Zheng\textsuperscript{1,2,$\dagger$},
Zehao~Hou\textsuperscript{3,$\dagger$},
and Rui~Xu\textsuperscript{1,2,$\dagger$},~\IEEEmembership{Senior~Member,~IEEE}\\[1ex]
{\footnotesize
\textsuperscript{1}Changsha University of Science and Technology, Changsha, China\\
\textsuperscript{2}School of Electrical and Information Engineering, Changsha University of Science and Technology, Changsha, China\\
\textsuperscript{3}University of Arkansas, Fayetteville, AR 72701, USA\\[0.5ex]
\textsuperscript{$\dagger$}These authors contributed equally to this work and share first authorship.\\
Corresponding author: Yu Zheng (e-mail: zhy9639@hotmail.com).\\
E-mail: ycong744@gmail.com (Cong Yao); zhou@uark.edu (Zehao Hou).\\
E-mail: gandiymaabyamb10@mails.tsinghua.edu (Rui Xu).
}}}
\maketitle
\begin{abstract}
Cross-material magnetic core loss prediction requires adapting to new material responses with limited measurements. Global feature alignment overlooks differences in source and target excitation coverage. This motivates selecting comparable samples and controlling each source measurement's alignment contribution. This paper proposes T-PCOT, a Transformer transfer method based on condition-supported partial optimal transport. Frequency, flux excursion, temperature, and waveform descriptors define a sparse graph of comparable operating conditions. A capacity-constrained transport plan distributes alignment across source measurements and leaves costly target mass unmatched. The plan is constructed from measured inputs before training and guides local waveform-feature consistency. An importance-weighted sampling scheme preserves both uniform source regression and the transport objective during minibatch optimization. Experiments on three material pairs and three target label budgets show that T-PCOT reduces mean relative error (MRE) and 95th-percentile error (P95) relative to Transformer-based multiple-kernel maximum mean discrepancy (T-MKMMD) in all nine comparisons. Macro MRE decreases from 12.60\% to 9.65\%, and macro P95 from 30.94\% to 25.06\%, corresponding to relative reductions of 23.41\% and 19.01\%. Controlled comparisons and ablations examine the roles of input processing, condition correspondence, and source capacity. The results demonstrate the predictive advantage of organizing cross-material alignment through operating-condition support over the compared global alignment method. The proposed method achieves state-of-the-art predictive performance in magnetic core loss transfer learning.
\end{abstract}
\section{Introduction}
\IEEEPARstart{M}{agnetic} core loss prediction supports efficient, compact power-converter design. Accurate models must capture the joint effects of frequency, flux excursion, waveform, temperature, and material properties. Measuring these combinations for each material is costly. Cross-material transfer learning addresses this cost by pretraining on a well-characterized source material and adapting with limited target measurements~\cite{dogariu2021}.

Source data may cover conditions rarely encountered by the target, while several target conditions may depend on the same few source measurements. Global feature-distribution alignment does not specify comparable sample pairs or their individual contributions, so reducing distribution discrepancy may align features across substantially different conditions. This raises the question: \emph{how can cross-material alignment use the operating conditions actually supported by source measurements?}

\begin{figure*}[!t]
\centering
\includegraphics[width=\textwidth]{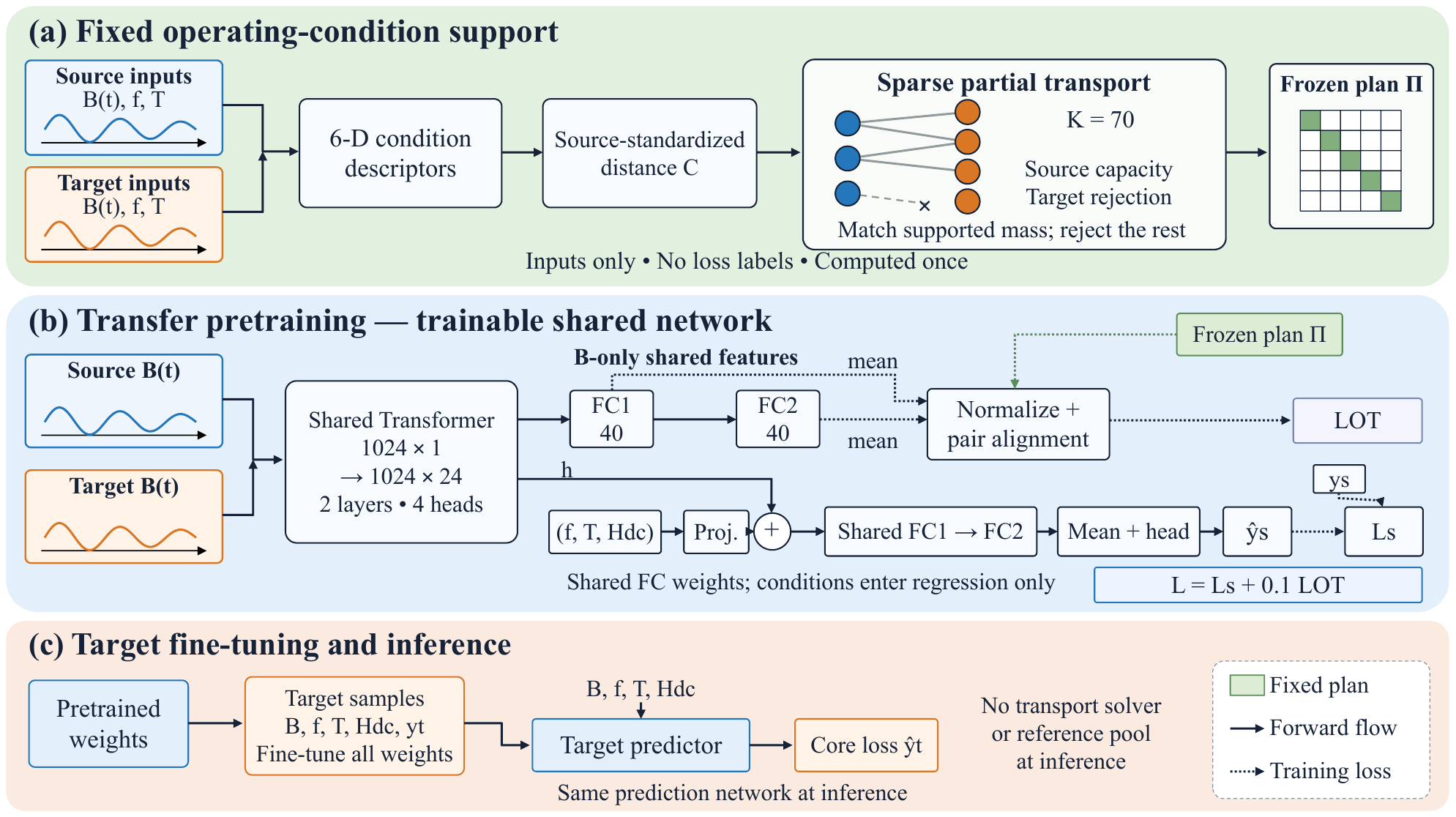}
\caption{Overview of \pcot{}. Measured inputs define a fixed transport plan through operating-condition neighborhoods, source capacities, and target rejection. During pretraining, this plan weights consistency between two waveform-feature layers, while source labels supervise loss regression. Target fine-tuning updates the shared predictor. Inference uses the waveform and operating conditions through the prediction path, including the fixed log-loss decoding.}
\label{fig:framework}
\end{figure*}

Operating conditions are central to core loss modeling, from Steinmetz formulations and their nonsinusoidal extensions~\cite{steinmetz,reinert,igse} to knowledge-aware neural networks~\cite{kann2024}. MagNet provides measurement infrastructure~\cite{howmagnet,whymagnet}, and MagNet-AI learns magnetic-material representations~\cite{magnetai}. CAHR-Net uses frequency, temperature, and waveform statistics to modulate hysteresis reconstruction for loss estimation~\cite{cahr}. Related engineering optimization also incorporates operating conditions, as in IHRSO for multi-UAV cooperative path planning under low-altitude winds~\cite{ihrso2026}. Here, operating conditions determine sample relationships for cross-material transfer pretraining.

MagLearn combines LSTM-based prediction with transfer and few-shot training through augmentation and alignment~\cite{maglearn2024}. Transformer-based multiple-kernel maximum mean discrepancy (T-MKMMD)~\cite{tmkmmd} combines a Transformer predictor with cross-domain feature alignment and subsequently fine-tunes the model on target measurements. Maximum mean discrepancy compares distributions through kernel mean embeddings~\cite{gretton}, and deep adaptation networks apply it to learned features~\cite{long}. Domain-adversarial training instead learns task-discriminative features that reduce domain distinguishability~\cite{dann2016}. T-MKMMD therefore provides a relevant comparison for transferring a core loss predictor using unlabeled target inputs. Its alignment objective, however, does not attach an explicit operating-condition constraint to each source--target interaction.

Our key idea is to use measured conditions to organize local representation alignment. Similar excitation conditions provide a basis for comparing waveform features, while target supervision accounts for the material-dependent loss response. This division matters because two materials can exhibit different losses under the same excitation. In transfer learning~\cite{pan}, a smaller feature discrepancy alone does not determine target prediction error when conditional relationships differ~\cite{bendavid}. T-PCOT uses input proximity to select and weight representation interactions; it learns target loss values through supervised adaptation.

Optimal transport makes these interactions explicit by assigning mass to sample pairs~\cite{courty,peyre}. DeepJDOT jointly aligns learned representations and label information through optimal transport~\cite{deepjdot2018}. Partial and unbalanced transport allow some mass to remain unmatched~\cite{chapel,chizat}. T-PCOT constructs a fixed sparse plan from measured inputs. Neighborhoods restrict interactions to nearby conditions, capacities limit each source measurement's contribution, and rejection leaves costly target mass unassigned. This plan weights local waveform-feature consistency independently of the evolving predictor.

Controlled experiments test condition correspondence by shuffling source associations, source reuse by removing capacities, and mass selection by maximizing feasible transport. Restricting source frequency at a fixed label count further examines incomplete condition support. These comparisons link the design to transport behavior and prediction errors.

The contributions are summarized as follows.
\begin{enumerate}
\item \textbf{Condition-driven sparse matching.} We introduce an operating-condition representation and sparse matching graph for cross-material core loss transfer. The graph uses frequency, flux excursion, temperature, and waveform shape to specify the source--target relationships eligible for alignment.
\item \textbf{Capacity-constrained partial transport.} We formulate capacity-constrained partial transport on this graph. Source capacities control the reuse of individual measurements, while unmatched target mass makes incomplete condition support explicit.
\item \textbf{Objective-preserving weighted training.} We derive an importance-weighted training scheme that preserves uniform source regression and the local transport objective. The method integrates into the existing Transformer pretraining and target-adaptation pipeline without adding a prediction module.
\item \textbf{Improved cross-material prediction.} Across three material pairs and three target label budgets, T-PCOT achieves lower mean relative error (MRE) and 95th-percentile error (P95) than T-MKMMD in all nine comparisons. Macro MRE and P95 decrease by 23.41\% and 19.01\%, respectively. The proposed method achieves state-of-the-art predictive performance in magnetic core loss transfer learning.
\end{enumerate}
Section~\ref{sec:method} presents T-PCOT. Section~\ref{sec:setup} describes the experimental setup, and Section~\ref{sec:results} evaluates predictive accuracy, transport behavior, and computational cost.
\section{Proposed Method}\label{sec:method}
\subsection{Problem Formulation}
Let $\mathcal D_s=\{(x_i^s,y_i^s)\}_{i=1}^{n_s}$ be the labeled source training set and $\mathcal U_t=\{x_j^t\}_{j=1}^{n_u}$ the unlabeled target pool used during pretraining. Each input contains a periodic flux-density waveform $B\in\R^{1024}$, frequency $f$, temperature $T$, and a dc-bias coordinate. The latter is zero for the available measurements. The positive response $y$ is volumetric core loss. A small labeled target set supports supervised adaptation; separate validation and test sets support model selection and evaluation.

T-PCOT separates two roles of operating conditions. In the prediction path, they help determine the material's loss response. In the alignment path, they decide which source and target waveform features should interact. Figure~\ref{fig:framework} shows this separation. It allows the matching relationships to be specified from measured inputs while the predictor learns material-dependent responses.

All methods share the Transformer backbone used for the T-MKMMD comparison. A pointwise projection maps the waveform to 24 channels, followed by sinusoidal positional encoding and two Transformer encoder layers~\cite{vaswani}. Each layer has four attention heads and a feedforward width of 40. A projected operating-condition vector is added to the encoded sequence before two shared fully connected blocks of width 40. Temporal mean pooling and a linear head produce $\widehat\ell_\theta(x)$, which is decoded as $\widehat y=10^{\widehat\ell_\theta+5}$.

The alignment path applies the same fully connected blocks to waveform encodings before adding the condition vector. Temporal averaging produces $z_\theta^{(1)},z_\theta^{(2)}\in\R^{40}$. These features share the regression weights. For T-PCOT, $\bar z=z/\max(\|z\|_2,10^{-8})$ normalizes each feature vector. Conditions therefore influence the choice and strength of pairwise interactions without directly satisfying the feature-consistency penalty through their own projection.

\subsection{Operating-Condition Description}
A useful matching descriptor must retain excitation scale while identifying equivalent representations of a periodic waveform. For a waveform of length $L=1024$, define its half excursion $A=(\max B-\min B)/2$, centered values $b_k=B_k-L^{-1}\sum_h B_h$, and normalized increments $d_k=|B_{k+1}-B_k|/A$, using circular indexing. We use
\begin{equation}\label{eq:descriptor}
\begin{split}
v(x)=\Bigg[&\log_{10}f,\ \log_{10}A,\ \frac{T-50}{40},\\
&\frac{\sqrt{L^{-1}\sum_k b_k^2}}{A},\ \log_{10}\!\left(L^{-1}\sum_k d_k^{1.5}\right),\ \frac{\operatorname{std}(d)}{\operatorname{mean}(d)}\Bigg].
\end{split}
\end{equation}
The first three components describe operating scale. The remaining components describe normalized waveform magnitude and slope variation. Circular shifts and sign reversal leave them unchanged, making the matching geometry insensitive to cycle origin and flux-axis direction. The fixed exponent 1.5 emphasizes waveform increments; it is a shape descriptor rather than a fitted material-loss exponent.

Descriptors are centered and scaled using source training means and standard deviations. A coordinate with source standard deviation below $10^{-8}$ is inactive. A target sample that differs from this constant source coordinate by more than $10^{-6}$ is excluded from matching; otherwise the coordinate is omitted. For normalized descriptors $u_i^s,u_j^t$ with $d_a$ active dimensions, the pairwise cost is
\begin{equation}\label{eq:cost}
C_{ij}=\frac{1}{d_a}\|u_i^s-u_j^t\|_2^2.
\end{equation}
Source-based scaling measures each discrepancy relative to the variation available in the source data. The descriptor summarizes excitation conditions, so proximity supplies a matching criterion rather than an equality constraint on the two materials' losses.

For each target input, the $K$ nearest source descriptors define the candidate edges $\mathcal E$. We choose the neighborhood scale from source inputs and the source--target pool sizes. Let $r_{\mathrm{cap}}=\lceil n_s/n_u\rceil$, $k_\tau=\lceil1.5r_{\mathrm{cap}}\rceil$, and $K=\min(n_s,\max(64,2k_\tau))$. The rejection price $\tau$ is the 95th percentile of the normalized squared distance between each source input and its $k_\tau$th other source neighbor. For $n_s=9000$ and $n_u=400$, these rules give $k_\tau=35$ and $K=70$. Neither measured losses nor test inputs enter this construction.

The quantity $r_{\mathrm{cap}}$ connects the neighborhood to the mass available for matching. A target column of mass $1/n_u$ needs at least $\lceil n_s/n_u\rceil$ source rows when each row supplies at most $1/n_s$. The factor 1.5 provides a margin around this minimum demand, and the candidate graph includes twice the resulting number of neighbors. Thus the graph provides alternative nearby sources when the closest source has already reached its capacity.

\subsection{Partial Transport}
Assign masses $a_i=1/n_s$ and $b_j=1/n_u$ to source and target samples. A nonnegative transport plan $\Pi$ weights candidate pairs, and $r_j$ represents target mass left unmatched. We solve
\begin{align}\label{eq:transport}
\min_{\Pi,r}\quad &\sum_{(i,j)\in\mathcal E}C_{ij}\Pi_{ij}+\tau\sum_j r_j,\\
\text{subject to}\quad &\sum_j\Pi_{ij}\leq a_i,\qquad
\sum_i\Pi_{ij}+r_j=b_j,\nonumber\\
&\Pi_{ij}=0\quad\text{for }(i,j)\notin\mathcal E.\nonumber
\end{align}
The graph controls which interactions are admissible. The row constraint limits the contribution of every source measurement. The column equality allocates each target sample's mass between matched sources and rejection. These decisions jointly determine where alignment occurs and how much mass it receives.

Eliminating $r_j$ gives the equivalent objective $\tau+\sum_{ij}(C_{ij}-\tau)\Pi_{ij}$ with upper bounds on both marginals. An edge below $\tau$ can reduce the objective, while an edge above $\tau$ is more expensive than rejection. Source capacity still couples the decisions across targets: two nearby target conditions may compete for the same source measurements. Once those sources reach capacity, either alternative neighbors supply mass or part of the target demand remains unmatched. If every candidate cost exceeds $\tau$, zero transport is optimal.

We solve the sparse linear program using the HiGHS dual-simplex backend in SciPy~\cite{scipy}. An input-identifier-based secondary objective resolves ties within the prescribed numerical tolerance. Row and column constraint violations are checked against $10^{-8}$. The resulting plan is computed once and remains fixed during network training. Keeping the geometry fixed prevents evolving prediction features from redefining the matching criterion.

Three quantities summarize the plan: total matched mass $M=\sum_{ij}\Pi_{ij}$, target coverage $\kappa_j=n_u\sum_i\Pi_{ij}$, and source utilization $\eta_i=n_s\sum_j\Pi_{ij}$. Equation~\eqref{eq:transport} ensures $M,\kappa_j,\eta_i\in[0,1]$. The rejected fraction is $1-M$. A target can receive positive mass while remaining partly uncovered, so matched sample counts and matched mass describe different aspects of source support. These quantities can be evaluated before training because they depend only on measured inputs.

\subsection{Training and Prediction}
The plan weights consistency between the two normalized waveform-feature layers:
\begin{equation}\label{eq:feature-distance}
D_\theta(i,j)=\frac12\sum_{h=1}^{2}\|\bar z_\theta^{(h)}(x_i^s)-\bar z_\theta^{(h)}(x_j^t)\|_2^2.
\end{equation}
With $\ell_i^s=\log_{10}y_i^s-5$, the pretraining objective is
\begin{equation}\label{eq:pretrain}
\mathcal L(\theta)=\sum_i a_i(\widehat\ell_\theta(x_i^s)-\ell_i^s)^2
+\lambda\sum_{ij}\Pi_{ij}D_\theta(i,j),\qquad\lambda=0.1.
\end{equation}
The first term uses all source labels uniformly. The second term, denoted $L_{\mathrm{OT}}=\sum_{ij}\Pi_{ij}D_\theta(i,j)$ in Fig.~\ref{fig:framework}, concentrates alignment on supported pairs. It retains the plan's absolute mass instead of dividing by $M$. With fixed feature distances, reduced supported mass therefore produces less alignment pressure.

Feature normalization makes the scale of this term explicit. Since each feature has norm at most one, $0\leq D_\theta(i,j)\leq4$ and $0\leq\lambda L_{\mathrm{OT}}\leq4\lambda M$. This bound concerns the objective value; predictive effects are evaluated experimentally. Source samples unused by transport still contribute to the regression loss, and unmatched target samples remain available for subsequent supervised adaptation.

Efficient sampling must preserve both terms of~\eqref{eq:pretrain}. Uniform sampling over all source--target pairs often draws zero-mass edges, whereas sampling only positive transport entries changes the marginal distribution of source supervision. We combine the two proposals. For $M>0$ and $\xi=0.5$,
\begin{align}\label{eq:proposal}
q_{ij}&=\frac{1-\xi}{n_sn_u}+\xi\frac{\Pi_{ij}}{M},\\
q_{s,i}&=\frac{1-\xi}{n_s}+\xi\frac{\sum_j\Pi_{ij}}{M}.\nonumber
\end{align}
For $m$ independent pair draws $(i_b,j_b)$, the minibatch objective is
\begin{align}\label{eq:estimator}
\widehat{\mathcal L}=\frac1m\sum_{b=1}^m\bigg[&\frac{a_{i_b}}{q_{s,i_b}}(\widehat\ell_\theta(x_{i_b}^s)-\ell_{i_b}^s)^2\\
&+\lambda\frac{\Pi_{i_bj_b}}{q_{i_bj_b}}D_\theta(i_b,j_b)\bigg].\nonumber
\end{align}
The source weight uses the marginal probability $q_{s,i}$ because regression depends only on the source draw. The alignment weight uses the joint probability $q_{ij}$ because feature consistency depends on both samples. Taking expectations yields $\sum_iq_{s,i}(a_i/q_{s,i})e_i^2=\sum_ia_ie_i^2$ and $\sum_{ij}q_{ij}(\Pi_{ij}/q_{ij})D_{ij}=\sum_{ij}\Pi_{ij}D_{ij}$. The arithmetic minibatch mean is therefore an unbiased estimator of the intended objective. The uniform component keeps every source probability positive; repeated indices remain valid independent draws. When $M=0$, sampling is uniform and the alignment term is zero.

After pretraining, supervised fine-tuning updates all prediction-network weights using target labels in the same log-loss space. Target validation selects the checkpoint with the lowest MRE, breaking ties by lower P95 and then earlier update. The transport plan remains unchanged throughout adaptation.

Inference executes only the regression path. It maps the waveform and operating conditions to a positive loss prediction without a neighbor search, transport solver, importance weights, or reference pool. The regression-only function is checked against the checkpoint's training interface before timing. T-PCOT thus places condition-aware interactions in training while retaining the original prediction interface.
\section{Experimental Setup}\label{sec:setup}
\subsection{Data and Evaluation Metrics}
We evaluate N87$\to$N27, 3C90$\to$3C94, and 77$\to$78 using the available MagNet MC1 measurements. These transfers cover three source--target material pairs with different operating-condition distributions. The release provides measurements at 25, 50, 70, and 90$^\circ$C with zero dc bias. The 70$^\circ$C level differs from the 75$^\circ$C level reported in the original T-MKMMD study. Both methods are evaluated on the same available measurements and data partition.

Near-duplicate inputs are grouped before splitting to keep closely related waveforms together. Candidate pairs have natural-log frequency differences at most $10^{-4}$, natural-log flux-excursion differences at most $10^{-3}$, and the same recorded temperature. Waveforms with relative RMS distance below $10^{-3}$ after circular shift and possible sign reversal form connected components. Representatives of these input-similarity groups are then used for the source and target partitions.

Each source material contributes 9000 training and 1000 validation labels. The target training sets contain 20, 100, or 400 labels, with 5, 25, or 100 corresponding validation labels. Thus the total target label budgets are 25, 125, and 500. The training pools are nested across budgets. During aligned pretraining, all 400 target training inputs are available without loss labels; target validation inputs are excluded. A disjoint test set contains 2000 target groups per material. These material pairs have also been used in earlier exploratory work; the present evaluation compares the methods under the common partition described here.

For each test measurement, relative error is $e=100|\widehat y-y|/y$. We report MRE and P95, both in percent. MRE describes average prediction accuracy, while P95 characterizes errors near the upper end of the distribution. Macro metrics give equal weight to each of the nine material--budget combinations. The geometric mean of paired T-PCOT/T-MKMMD error ratios provides a complementary summary of relative changes across conditions. Frequency and amplitude slices use medians of the 400-input target training pool; temperature slices use the recorded levels.

\subsection{Comparison Methods}
The main comparison uses T-MKMMD with the same Transformer backbone and target fine-tuning procedure. Its Gaussian kernel scales are $\{0.25,0.5,1,2,4\}$ times the median nonzero squared distance of the combined source--target batch. Softmax kernel weights are shared by the two feature layers and updated by Adam ascent with learning rate 0.01. The biased V-statistic is computed within each eight-sample microbatch. Four-microbatch accumulation yields 32 samples per network update, while retaining the eight-sample kernel estimates. Target-waveform augmentation consists of circular shift, sign reversal, and amplitude scaling in $[0.9,1.1]$; alignment features remain unnormalized.

A matched-clean MKMMD control uses clean target waveforms and the same feature normalization as T-PCOT. It retains the multiple-kernel discrepancy objective and runs at the 125-label budget on all three pairs. This comparison tests whether the advantage of T-PCOT persists when input processing and feature normalization are matched. Because these two settings change together, the control evaluates their combined effect.

Three ablations use N87$\to$N27 with 125 target labels. \emph{Shuffled} permutes the transport plan's source rows, preserving total mass, target column sums, and the distribution of source row sums while changing condition correspondence. \emph{No capacity} removes the source row constraint and retains the cost, candidate graph, and rejection price. \emph{Maximum mass} first maximizes feasible transported mass on the same graph under both capacities, then minimizes condition cost at that mass.

The source-coverage experiment restricts N87 inputs to frequencies no higher than the median of the source input groups, 141250~Hz. Both methods retain 9000 source training and 1000 source validation labels. Target indices and label budgets remain unchanged, so the experiment isolates a change in source coverage at the same nominal measurement budget.

\subsection{Implementation Details}
Table~\ref{tab:protocol} summarizes the common training settings. Adam~\cite{adam} uses cosine learning-rate decay~\cite{sgdr} without restarts. Dropout is 0.1 and gradients are clipped at norm 5. Waveforms are divided by 0.3; frequency and temperature inputs are transformed as $(\log_{10}f-5)/0.5$ and $(T-50)/40$; labels use $\log_{10}y-5$. All prediction networks share these transformations and the same update budgets. Validation selects checkpoints before test evaluation, and the same test observations are used for each paired comparison.

\begin{table}[!tb]
\caption{Training and Evaluation Settings}\label{tab:protocol}
\centering\small
\begin{tabular}{@{}ll@{}}\toprule
Item & Setting \\\midrule
Source labels & 9000 training + 1000 validation \\
Target total labels & 25 / 125 / 500 \\
Target train + validation & 20+5 / 100+25 / 400+100 \\
Unlabeled target inputs & 400 during aligned pretraining \\
Test samples & 2000 groups per target \\
Pretraining & 6000 updates, learning rate 0.001 \\
Fine-tuning & 9000 updates, learning rate 0.0003 \\
Optimizer and batch & Adam; $8\times4$ accumulation \\
Schedule & Cosine decay to 5\% of initial rate \\
Alignment strength & 0.1 \\
Checkpoint interval & 250 updates \\\bottomrule
\end{tabular}
\end{table}

Training uses shared RTX 4090 GPUs with a 4~GiB allocator limit per worker. Computational costs are measured under this shared-device setting. Inference timing uses 20 warm-up passes followed by 100 timed CUDA forward passes for batch sizes one and eight. The reported latency is the median time per batch.
\section{Results and Discussion}\label{sec:results}
We first evaluate prediction accuracy against T-MKMMD, then examine the alignment settings and individual transport decisions. Source-coverage and error analyses show how the method behaves across operating conditions. Computational measurements complete the comparison.

\subsection{Prediction Accuracy}
\begin{figure*}[!tb]
\centering\includegraphics[width=\textwidth]{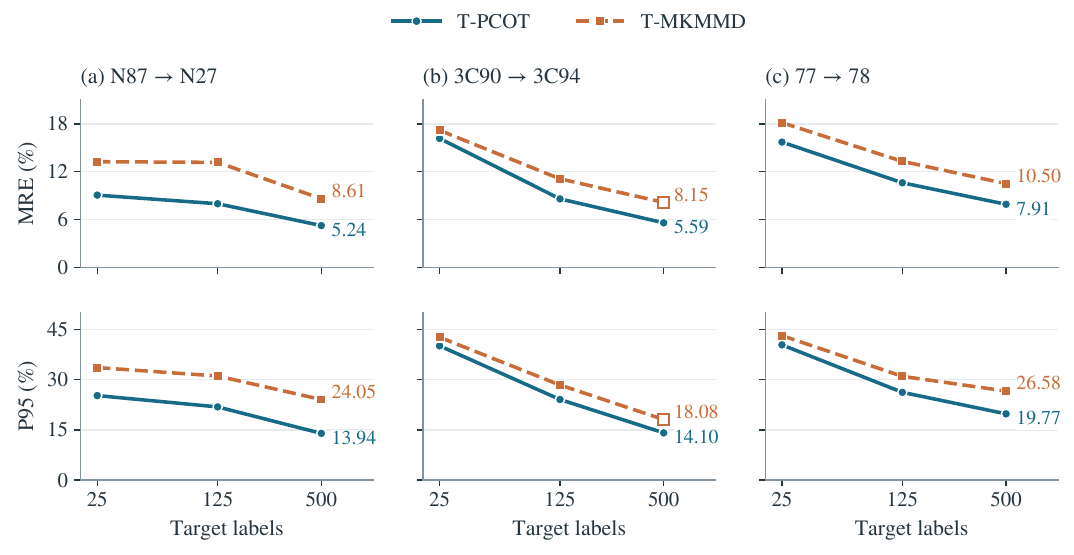}
\caption{Prediction errors across target label budgets, using 2000 target test groups. Open orange squares mark the 3C90$\to$3C94, 500-label T-MKMMD run with an unresolved convergence diagnostic at 9000 updates; its checkpoint is selected by validation.}
\label{fig:main}
\end{figure*}

T-PCOT achieves lower MRE and P95 than T-MKMMD in all nine material--budget combinations, as shown in Table~\ref{tab:main} and Fig.~\ref{fig:main}. Macro MRE decreases from 12.60\% to 9.65\%, a relative reduction of 23.41\%. Macro P95 decreases from 30.94\% to 25.06\%, a relative reduction of 19.01\%. The geometric means of paired MRE and P95 ratios are 0.739 and 0.784, respectively. The advantage therefore appears in both average prediction error and the upper portion of the error distribution.

\begin{table*}[!tb]
\caption{Prediction Errors Across Materials and Label Budgets. Errors are in percent; $\Delta$ MRE is relative to T-MKMMD.}\label{tab:main}
\centering\small\begin{tabular}{@{}lrrrrrr@{}}
\toprule
Pair & Labels & T-MKMMD MRE & T-PCOT MRE & T-MKMMD P95 & T-PCOT P95 & $\Delta$ MRE (\%) \\
\midrule
N87$\to$N27 & 25 & 13.25 & 9.06 & 33.59 & 25.22 & -31.6 \\
N87$\to$N27 & 125 & 13.17 & 7.98 & 31.06 & 21.85 & -39.4 \\
N87$\to$N27 & 500 & 8.61 & 5.24 & 24.05 & 13.94 & -39.2 \\
3C90$\to$3C94 & 25 & 17.18 & 16.17 & 42.63 & 40.07 & -5.9 \\
3C90$\to$3C94 & 125 & 11.10 & 8.58 & 28.38 & 24.05 & -22.7 \\
3C90$\to$3C94 & 500 & 8.15 & 5.59 & 18.08 & 14.10 & -31.4 \\
77$\to$78 & 25 & 18.14 & 15.71 & 43.14 & 40.37 & -13.4 \\
77$\to$78 & 125 & 13.29 & 10.61 & 30.98 & 26.19 & -20.2 \\
77$\to$78 & 500 & 10.50 & 7.91 & 26.58 & 19.77 & -24.7 \\
\bottomrule
\end{tabular}
\end{table*}

The size of the gain depends on the material pair. For N87$\to$N27, relative MRE reductions range from 31.6\% to 39.4\%. At the 125-label budget, MRE falls from 13.17\% to 7.98\%, while P95 falls from 31.06\% to 21.85\%. Improvements on 3C90$\to$3C94 are smaller at 25 labels, where MRE changes from 17.18\% to 16.17\%, and grow at the larger budgets. On 77$\to$78, the MRE reduction increases from 13.4\% at 25 labels to 24.7\% at 500 labels. These differences show why both material pairs and label budgets are needed to assess cross-material transfer.

Increasing the target label budget improves both metrics for both methods on every material pair. T-PCOT also reaches some accuracy levels with fewer labels: on N27, its 125-label MRE of 7.98\% is below the 8.61\% achieved by T-MKMMD at 500 labels. This comparison uses the measured budget points directly. The 3C94, 500-label T-MKMMD condition retains an unresolved convergence diagnostic at the common 9000-update limit. Its validation-selected checkpoint is included in the comparison, so that condition measures performance at the specified training budget.

\subsection{Alignment Settings}
The matched-clean control examines whether input processing explains the main comparison. It uses the same clean target waveforms and feature normalization as T-PCOT while retaining MKMMD. Table~\ref{tab:controls} shows that T-PCOT has lower MRE and P95 on all three pairs under this matched setting.

\begin{table}[!tb]
\caption{Alignment Comparison at 125 Target Labels}\label{tab:controls}
\centering\small\begin{tabular}{@{}llrr@{}}
\toprule
Target & Method & MRE (\%) & P95 (\%) \\
\midrule
N27 & T-MKMMD & 13.17 & 31.06 \\
N27 & T-PCOT & 7.98 & 21.85 \\
N27 & Matched-clean MKMMD & 12.74 & 31.14 \\
3C94 & T-MKMMD & 11.10 & 28.38 \\
3C94 & T-PCOT & 8.58 & 24.05 \\
3C94 & Matched-clean MKMMD & 15.46 & 35.47 \\
78 & T-MKMMD & 13.29 & 30.98 \\
78 & T-PCOT & 10.61 & 26.19 \\
78 & Matched-clean MKMMD & 14.14 & 35.15 \\
\bottomrule
\end{tabular}
\end{table}

For N27, the control gives 12.74\% MRE, compared with 7.98\% for T-PCOT. The corresponding comparisons are 15.46\% versus 8.58\% on 3C94 and 14.14\% versus 10.61\% on 78. Matching input processing therefore preserves the advantage of condition-supported transport over kernel-discrepancy alignment. The control itself improves on the original MKMMD MRE for N27 but worsens it for the other two materials. Clean waveforms and normalized features do not provide a uniform gain by themselves; their effect depends on the alignment objective and material pair.

\subsection{Ablation Study}
\begin{figure*}[!tb]
\centering\includegraphics[width=\textwidth]{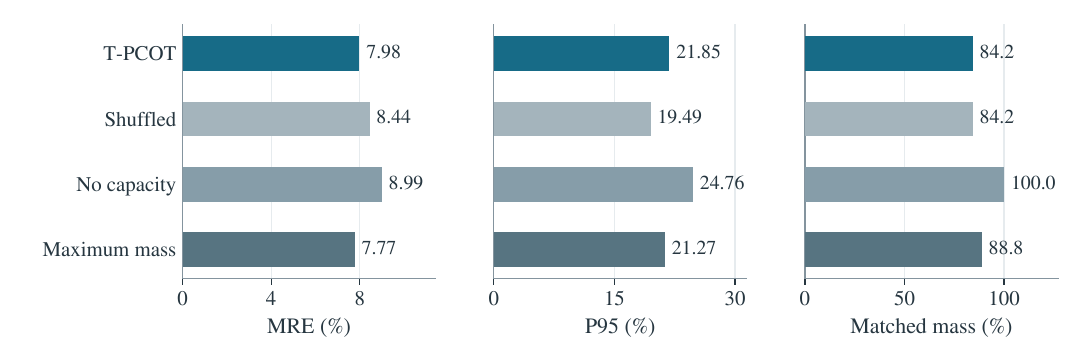}
\caption{Transport ablations on N87$\to$N27 with 125 target labels. MRE and P95 use the same 2000 target test groups. Matched mass is the fraction of demand assigned across 400 target inputs. The variants change source associations, source capacities, or feasible matching mass.}
\label{fig:mechanism}
\end{figure*}

The ablations isolate three decisions in the transport plan: which conditions interact, how strongly a source measurement can contribute, and how much feasible mass is accepted. Figure~\ref{fig:mechanism} and Table~\ref{tab:ablation} report their effects on N87$\to$N27 at the 125-label budget.

\begin{table*}[!tb]
\caption{Ablation Results and Transport Statistics}\label{tab:ablation}
\centering\small\begin{tabular}{@{}lrrrrr@{}}
\toprule
Variant & MRE (\%) & P95 (\%) & $M$ (\%) & Sources $>0$ & Mean $C$ \\
\midrule
T-PCOT & 7.98 & 21.85 & 84.20 & 7593 & 0.053 \\
Shuffled & 8.44 & 19.49 & 84.20 & 7593 & 1.929 \\
No capacity & 8.99 & 24.76 & 100.00 & 388 & 0.007 \\
Maximum mass & 7.77 & 21.27 & 88.78 & 7999 & 0.068 \\
\bottomrule
\end{tabular}
\end{table*}

Shuffling source associations preserves the plan's total mass and marginal distributions while raising mean condition cost from 0.0535 to 1.9287. MRE increases from 7.98\% to 8.44\%, showing an average-error benefit from the original condition correspondence. P95 moves in the opposite direction, decreasing from 21.85\% to 19.49\%. The two metrics reveal different consequences of the pairing structure: preserving nearby conditions helps the mean in this comparison, while the upper-tail response also depends on how the learned representations change.

Source capacity has a clearer effect on both metrics. Removing it increases MRE to 8.99\% and P95 to 24.76\%. Almost unit transport mass then concentrates on 388 source measurements, compared with 7593 under T-PCOT. Maximum normalized source utilization rises from one to 45. The capacity constraint distributes alignment across a much broader set of measurements and produces lower errors in this ablation. Because capacity also affects feasible mass and rejection, the experiment evaluates the complete consequence of removing that constraint.

Maximum-feasible matching increases transported mass from 84.20\% to 88.78\%, with mean condition cost increasing from 0.0535 to 0.0680. It also improves MRE to 7.77\% and P95 to 21.27\%. Thus accepting additional, more costly matches is beneficial in this case. Cost-based rejection gives an explicit way to regulate accepted mass, but its default level is not the most accurate among these variants. The result identifies mass selection as an avenue for improving the method while preserving the observed benefit of source capacity.

\subsection{Source Coverage}
\begin{figure*}[!tb]
\centering\includegraphics[width=\textwidth]{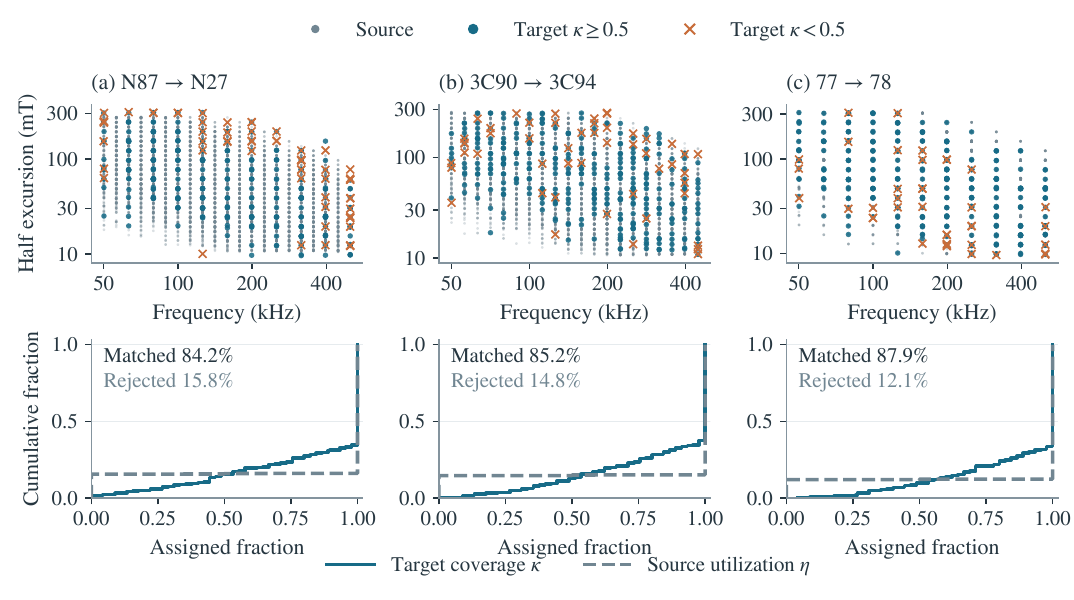}
\caption{Operating-condition support across material pairs. Top: frequency--flux projections of 9000 source and 400 target inputs; orange crosses mark target coverage below 0.5. Bottom: cumulative distributions of target coverage $\kappa$ and source utilization $\eta$. The transport plan uses all six condition descriptors.}
\label{fig:support}
\end{figure*}

Source support determines which local interactions can enter training. Figure~\ref{fig:support} shows frequency--flux projections of the source and target inputs and the resulting distributions of target coverage and source utilization. The plan uses all six descriptors, so the projections visualize only part of the matching geometry.

\begin{table*}[!tb]
\caption{Source Support and Transport Preparation Cost}\label{tab:support}
\centering\small\begin{tabular}{@{}lrrrrr@{}}
\toprule
Pair & $\tau$ & $M$ (\%) & Targets $>0$ & Sources $>0$ & Build (s) \\
\midrule
N87$\to$N27 & 0.2082 & 84.20 & 393 & 7593 & 11.08 \\
3C90$\to$3C94 & 0.2107 & 85.19 & 399 & 7685 & 11.45 \\
77$\to$78 & 0.1977 & 87.91 & 400 & 7925 & 10.49 \\
N87$\to$N27 (low $f$) & 0.2226 & 53.57 & 227 & 4860 & 3.12 \\
\bottomrule
\end{tabular}
\end{table*}

Matched masses are 84.20\%, 85.19\%, and 87.91\% for N87$\to$N27, 3C90$\to$3C94, and 77$\to$78. The numbers of target inputs receiving positive mass are 393, 399, and 400 out of 400; the corresponding source counts are 7593, 7685, and 7925. On 77$\to$78, every target input receives some mass, yet 12.09\% of the prescribed total remains unmatched. This illustrates the distinction between reaching a target condition and supplying its full demand under finite source capacity.

Restricting source frequency makes the coverage change more pronounced. The number of source labels is unchanged, but matched mass on N87$\to$N27 falls from 84.20\% to 53.57\%, and only 227 target inputs receive positive mass. The source-derived rejection price changes from 0.2082 to 0.2226 after restandardization. The lower coverage therefore occurs under the same threshold-construction rule.

\begin{figure*}[!tb]
\centering\includegraphics[width=\textwidth]{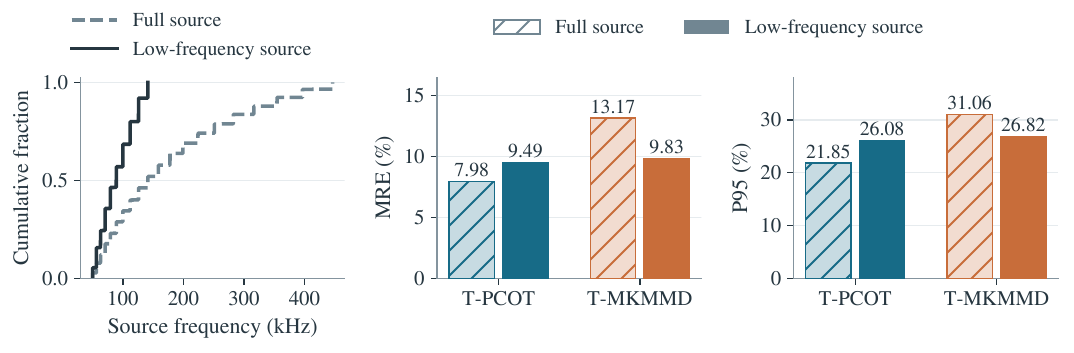}
\caption{Effect of restricting source frequency on N87$\to$N27 at 125 target labels. Restricted inputs satisfy $f\leq141250$~Hz; both regimes retain 9000 training and 1000 validation labels. The source CDF uses the 9000 training inputs. Error summaries use the same 2000 target test groups; hatched and solid bars denote full and restricted coverage.}
\label{fig:stress}
\end{figure*}

Figure~\ref{fig:stress} shows the associated prediction errors. T-PCOT MRE/P95 increase from 7.98/21.85\% to 9.49/26.08\%. T-MKMMD obtains 9.83/26.82\% with the restricted source, leaving a smaller advantage for T-PCOT than under full coverage. T-MKMMD itself improves when the source range is restricted, indicating that broader source coverage does not automatically improve global distribution alignment.

The coverage experiment also reveals a sensitivity in extreme predictions. Maximum T-PCOT error rises to 229.78\%, compared with 50.99\% for T-MKMMD. Increased rejected mass makes the loss of input support visible, but it does not calibrate prediction risk. Managing severe individual errors under restricted source coverage remains a distinct objective from improving MRE and P95.

\subsection{Error Analysis}
\begin{figure*}[!tb]
\centering\includegraphics[width=\textwidth]{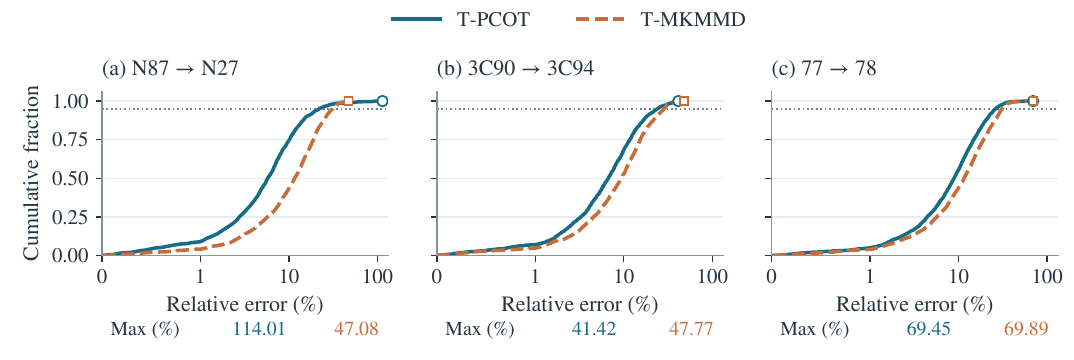}
\caption{Relative-error distributions at 125 target labels, using the same 2000 test groups per material. The dotted line marks cumulative fraction 0.95. The axis is linear near zero and logarithmic above 1\%. Hollow endpoints and color-matched values identify the maximum errors.}
\label{fig:cdf}
\end{figure*}

Figure~\ref{fig:cdf} compares the error distributions at 125 labels. Their shapes explain why MRE, P95, and maximum error should be distinguished. T-PCOT improves MRE and P95 on all three pairs, but the largest error on N27 is 114.01\%, compared with 47.08\% for T-MKMMD. Across all nine main conditions, maximum error favors T-PCOT in seven. Most of the distribution and its extreme endpoint therefore need separate interpretation.

\begin{figure*}[!tb]
\centering\includegraphics[width=\textwidth]{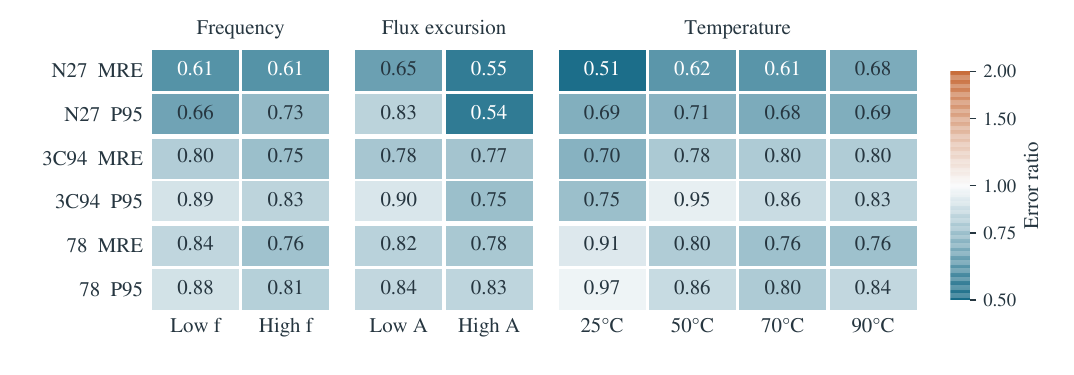}
\caption{T-PCOT/T-MKMMD error ratios at 125 labels. Ratios below one favor T-PCOT; $A$ denotes flux half excursion. Text and color-bar ticks give raw ratios; color uses a logarithmic scale centered at one. Slice counts in column order are N27: 934/1066/1033/967/483/509/510/498; 3C94: 971/1029/1040/960/487/543/459/511; 78: 998/1002/927/1073/473/495/512/520.}
\label{fig:slices}
\end{figure*}

Operating-condition slices provide a more detailed view of the improvements. Figure~\ref{fig:slices} reports T-PCOT/T-MKMMD error ratios for frequency, flux half excursion, and temperature. Both metrics are lower for T-PCOT in all 24 material--slice combinations. The improvement therefore extends across the examined ranges rather than arising from one frequency or amplitude partition.

Each slice compares the methods on identical observations. Frequency and amplitude partitions use training-input medians, and temperature partitions use measured levels. These overlapping views locate accuracy gains along different operating coordinates of the same test set.

\subsection{Computational Cost}

\begin{figure*}[!tb]
\centering\includegraphics[width=\textwidth]{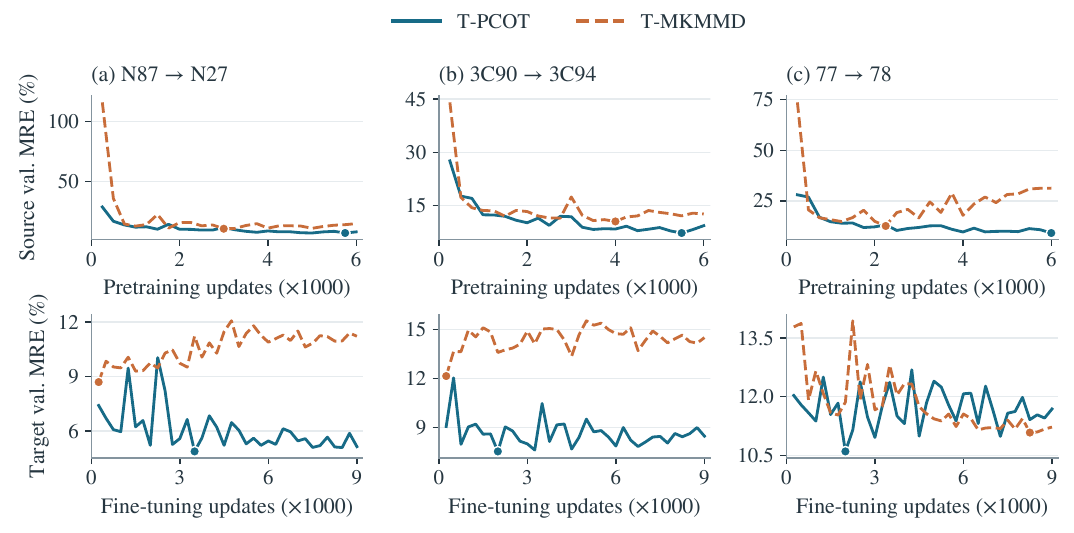}
\caption{Validation MRE during source pretraining and target fine-tuning with 125 target labels. The upper row uses source validation measurements and the lower row uses target validation measurements. Curves connect the recorded checkpoint evaluations; dots mark the validation-selected checkpoints.}
\label{fig:training}
\end{figure*}

Figure~\ref{fig:training} shows source-pretraining and target-fine-tuning validation trajectories. At 125 labels, the selected target checkpoints occur at updates 3500/2000/2000 for T-PCOT and 250/250/8250 for T-MKMMD on N27/3C94/78. Validation performance can deteriorate after its minimum, so selecting a checkpoint is consequential even when the total update budget is shared.

\begin{figure}[!tb]
\centering\includegraphics[width=\columnwidth]{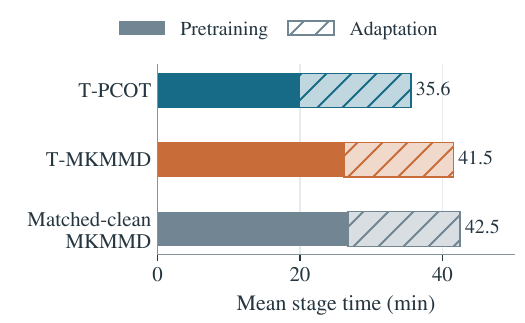}
\caption{Mean measured training time across three material pairs on shared RTX 4090 GPUs. Target adaptation uses 125 labels. Solid and hatched portions indicate pretraining and adaptation; labels give their total mean time.}
\label{fig:cost}
\end{figure}

\begin{table}[!tb]
\caption{Training Cost and Prediction Parameters}\label{tab:costs}
\centering\small\begin{tabular}{@{}lrrrr@{}}
\toprule
Method & Params. & Pre (s) & Tune (s) & MiB \\
\midrule
T-PCOT & 12385 & 1193.9 & 940.4 & 1486 \\
T-MKMMD & 12385 & 1571.6 & 920.0 & 1486 \\
Matched-clean MKMMD & 12385 & 1604.0 & 944.3 & 1486 \\
\bottomrule
\end{tabular}
\end{table}

Table~\ref{tab:costs} and Fig.~\ref{fig:cost} report mean stage times across the three material pairs using the 125-label fine-tuning runs. T-PCOT pretraining takes 1193.9~s, compared with 1571.6~s for T-MKMMD. Fine-tuning takes 940.4~s and 920.0~s, respectively. Both methods use approximately 1486~MiB of peak allocated memory. Input description, graph construction, and transport solution require a one-time 10.49--11.45~s for the full-source cases, excluding data loading. These measurements compare the complete training procedures on shared GPUs.

Both prediction networks have 12,385 parameters. Median batch-1 latency is 2.542~ms for T-PCOT and 2.541~ms for T-MKMMD; batch-8 latency is 3.704~ms and 3.706~ms, respectively. The regression-only function reproduces its checkpoint's training-interface output with zero maximum absolute log-prediction difference.
\section{Conclusion}\label{sec:conclusion}
This paper presented T-PCOT for magnetic core loss transfer learning across materials with limited target measurements. Its central design is to organize alignment through operating-condition support. A sparse input-based graph identifies comparable conditions, source capacities distribute their contribution, and partial transport permits expensive target mass to remain unmatched. Importance-weighted sampling combines these local interactions with uniform source supervision, while target adaptation and inference retain the shared Transformer predictor.

Across three material pairs and three label budgets, T-PCOT achieves lower MRE and P95 than T-MKMMD in all nine comparisons. Macro MRE decreases from 12.60\% to 9.65\%, and macro P95 from 30.94\% to 25.06\%. The matched-input comparison and capacity ablation further connect this advantage to the organization of alignment. The results identify two directions for further development: adapting the accepted transport mass to predictive needs and controlling extreme errors when source condition coverage contracts. These findings position operating-condition support as a useful basis for designing cross-material alignment in core loss prediction.
\label{sec:conclusion-end}

\label{sec:body-end}
\clearpage
\bibliographystyle{IEEEtran}
\bibliography{references}
\end{document}